\documentclass[letterpaper, 10 pt, conference]{ieeeconf}  % Comment this line out if you need a4paper

\usepackage{graphics} % for pdf, bitmapped graphics files
\usepackage{epsfig} % for postscript graphics files
\usepackage{amssymb}  % assumes amsmath package installed
\usepackage{amsmath} % assumes amsmath package installed
\usepackage{multirow}
\usepackage{booktabs}
\usepackage{makecell}
\usepackage{graphicx} % 用于插入图片
\usepackage{caption}  % 用于图片标题设置
\usepackage{cuted}
\usepackage{pifont}
\usepackage[pagebackref,breaklinks,colorlinks]{hyperref}

\title{\LARGE \bf
GarmentPile++: Affordance-Driven Cluttered Garments Retrieval with Vision-Language Reasoning
}

\vspace{-3mm}
\author{
\authorblockN{Mingleyang Li$^{1*}$, Yuran Wang$^{1*}$, Yue Chen$^{1}$, Tianxing Chen$^{2}$}
\authorblockN{Jiaqi Liang$^{1}$, Zishun Shen$^{1}$, Haoran Lu$^{3}$, Ruihai Wu$^{1\dagger}$, Hao Dong$^{1\dagger}$}
\authorblockA{Peking University$^{1}$, The University of Hong Kong$^{2}$, Northwestern University$^{3}$\\
$^*$Equal contribution \quad $^\dagger$Corresponding author\\
}
}

\begin{document}

\maketitle
\thispagestyle{empty}
\pagestyle{empty}

%%%%%%%%%%%%%%%%%%%%%%%%%%%%%%%%%%%%%%%%%%%%%%%%%%%%%%%%%%%%%%%%%%%%%%%%%%%%%%%%

\begin{strip}
\vspace{-12mm}
    \centering
    \includegraphics[width=\linewidth]{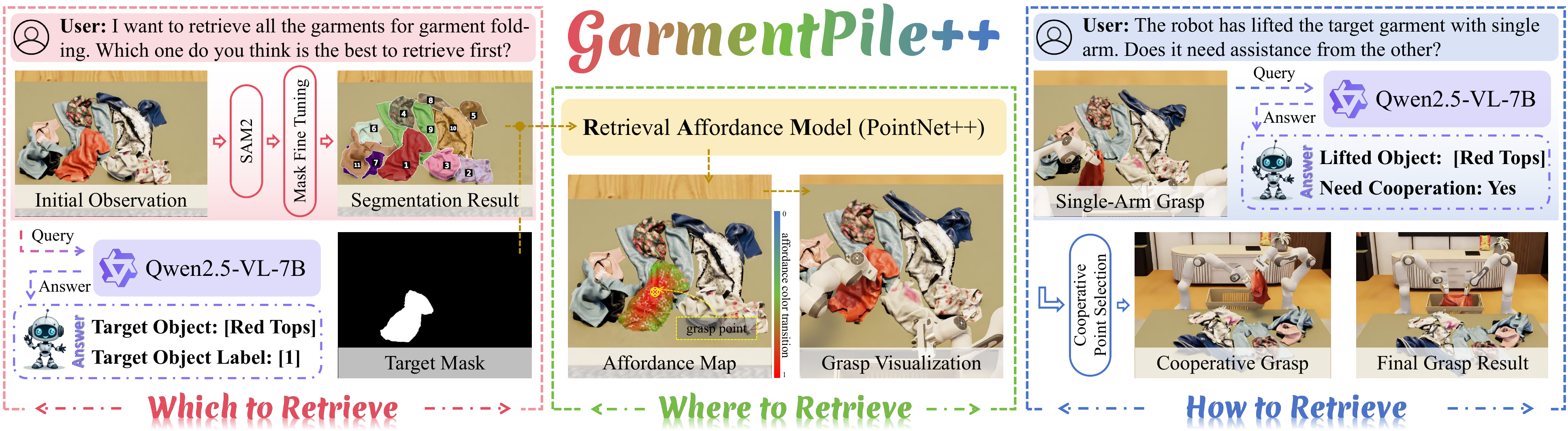}
    \vspace{-5mm}
    \captionof{figure}{\textbf{Overview.} GarmentPile++ mainly includes three stages: 1) \textbf{Which to Retrieve.} Leveraging SAM2 (with optionally triggered mask fine tuning), masks of garments are obtained to provide visual cues for aiding VLM-based reasoning on the optimal garment to retrieve under task constraints. 2) \textbf{Where to Retrieve.} The Retrieval Affordance Model infers optimal grasp points for the target garment, maximizing single-arm retrieval feasibility while ensuring the garment’s safety and cleanliness. 3) \textbf{How to Retrieve.} For large/long garments, single-arm retrieval may be unfeasible. The single-arm grasping condition is therefore fed to VLM to determine whether dual-arm cooperation is required, ensuring smooth garment retrieval.}  
    \label{fig:teaser}
\vspace{-5mm}
\end{strip}

\begin{abstract}
Garment manipulation has attracted increasing attention due to its critical role in home-assistant robotics. However, the majority of existing garment manipulation works assume an initial state consisting of only one garment, while piled garments are far more common in real-world settings. To bridge this gap, we propose a novel garment retrieval pipeline that can not only follow language instruction to execute safe and clean retrieval but also guarantee exactly one garment is retrieved per attempt, establishing a robust foundation for the execution of downstream tasks (e.g., folding, hanging, wearing). Our pipeline seamlessly integrates vision-language reasoning with visual affordance perception, fully leveraging the high-level reasoning and planning capabilities of VLMs alongside the generalization power of visual affordance for low-level actions. To enhance the VLM's comprehensive awareness of each garment's state within a garment pile, we employ visual segmentation model (SAM2) to execute object segmentation on the garment pile for aiding VLM-based reasoning with sufficient visual cues. A mask fine-tuning mechanism is further integrated to address scenarios where the initial segmentation results are suboptimal. In addition, a dual-arm cooperation framework is deployed to address cases involving large or long garments, as well as excessive garment sagging caused by incorrect grasping point determination, both of which are strenuous for a single arm to handle. The effectiveness of our pipeline are consistently demonstrated across diverse tasks and varying scenarios in both real-world and simulation environments. Project page: \href{https://garmentpile2.github.io/}{https://garmentpile2.github.io/}.

\end{abstract}
\section{INTRODUCTION}
As a key aspect of general home-assistant robot capabilities, garment manipulation is highly challenging because of its near-infinite space of self-deformable states and complex kinematic and dynamic properties. These properties become even more pronounced in cluttered scenarios, where garments are prone to entanglement with more complicated states, making successful retrieval challenging. Although significant progress has been made in garment manipulation including folding~\cite{xue2023unifolding, chen2025metafold}, flinging~\cite{ha2022flingbot, wang2025dexgarment}, hanging~\cite{wu2024unigarmentmanip, chen2025robohanger}, or dressing humans~\cite{sun2024force, kotsovolis2024model}, most of the existing work assumes the manipulation of a single garment. However, garment piles are far more common in everyday environments, significantly limiting the real-world deployability of these approaches. Therefore, the ability to retrieve individual garment from a pile is of critical importance.

A recent work, GarmentPile~\cite{wu2025garmentpile}, presents a promising approach for cluttered garment manipulation using a per-point affordance model to predict retrieval actionability. Furthermore, For scenarios where the overly complex cluttered state of garments leaves no suitable manipulation region, an additional adaptation model is trained to perform simple adjustments to the initial garment pile, boosting the performance of method. 
However, due to the specificity of their task scenarios (e.g., placing garments into washing machine), their approach doesn't consider a dedicated mechanism for single garment retrieval, often resulting in the retrieval of multiple garments simultaneously, which in turn limits the applicability of some downstream tasks. 
Moreover, due to its reliance on visual affordance alone, the method lacks integration with language, significantly constraining its flexibility. 
Furthermore, The garment retrieval process is based solely on a single arm, which struggles to handle large/long garments, therefore compromises the robustness and overall performance of the method in various types of garments.

To overcome these limitations, we propose GarmentPile++, a novel pipeline which not only leverages the high-level reasoning capabilities of VLMs, but also exploits the low-level manipulation accuracy and generalization of visual affordance models. 
As shown in Fig.~\ref{fig:teaser}, we structure the process into three stages: 
% The first stage focuses on retrieval target selection (i.e. \textbf{Which to Retrieve}):
\textbf{(1) Which to Retrieve:}
By integrating visual segmentation model (SAM2~\cite{ravi2024sam}) with VLM-based reasoning, we identify the garment which is suitable for retrieval under task constraints. Furthermore, to address the potential inaccurate segmentation caused by heavy occlusion and color similarity in garment pile, we introduce a mask fine-tuning procedure, which utilizes SAM2 video tracking and point-prompt segmentation to get better segmentation results.
\textbf{(2) Where to Retrieve:}
To maximize the feasibility of single-arm retrieval while ensuring garment safety and cleanliness, the Retrieval Affordance Model is trained to execute the garment grasping, in which the affordance representation is enhanced by incorporating segmentation mask feature into the garment pile's point cloud, enabling seamless integration into a language-guided manipulation pipeline. 
\textbf{(3) How to Retrieve:}
The dual-arm system is deployed for handling different garments and boosting the robustness of pipeline. After master-arm grasp and lift, the current scene situation will be reasoned by VLM for deciding whether to trigger dual-arm cooperation or not. If triggered, a tracking selection procedure will be used for selecting cooperation point for slave-arm grasp. What's more, the grasp of more than one garments will be detected by VLM to terminate current retrieval and continue next attempt.

We design two typical task scenarios encountered in daily life: open-boundary and closed-boundary. Within these scenarios, we define two retrieval tasks: 1) retrieve garments sequentially, 2) retrieve specific garment. These settings collectively cover nearly all practical cases of garment pile retrieval in everyday environments. Extensive experiments demonstrate the effectiveness of our pipeline.

In summary, our main contributions are as follows:
\begin{itemize}
    \item we propose a novel pipeline that follows language instruction to execute safe garment retrieval while ensuring exactly one garment is retrieved per attempt, establishing a robust foundation for downstream tasks.
    \item Our pipeline seamlessly combines vision-language reasoning with visual affordance, fully exploiting the high-level reasoning and planning capabilities of VLMs alongside the generalization power of affordance for low-level grasping.
    \item Through extensive experiments in both real-world and simulated environments, we demonstrate the effectiveness of our method by successfully tackling two distinct tasks across the two designed scenarios.
\end{itemize}

\section{RELATED WORK}

\subsection{Deformable Object and Garment Manipulation}
Deformable object manipulation poses significant challenges due to complex dynamics and high-dimensional state spaces~\cite{chen2025benchmarking}. Compared with manipulating 1D cables and ropes~\cite{seita2021learning, wu2019learning, zhang2024adaptigraph} or square-shaped fabrics~\cite{ganapathi2021learning, lee2020learning, wu2023learning, weng2022fabricflownet}, garments with diverse categories, shapes, and deformations are significantly more challenging to handle. Current research primarily focuses on manipulating single garments, proposing methods for folding~\cite{avigal2022speedfolding,xue2023unifolding}, unfolding~\cite{canberk2022cloth, ha2022flingbot, li2015regrasping}, hanging~\cite{chen2023learning, Wu_2024_CVPR}, and dressing~\cite{kotsovolis2024model,sun2024force, Wang2023One}. Recent advancements in simulation platforms like GarmentLab~\cite{lu2024garmentlab} and DexGarmentLab~\cite{wang2025dexgarment} have enabled systematic evaluation of garment manipulation techniques. However, most studies focus on isolated garments in structured environments. In this paper, we address the more challenging problem of cluttered garment manipulation.

\subsection{Cluttered Environment Manipulation}

Robotic manipulation in cluttered environments is a fundamental capability for tasks such as grasping~\cite{wang2021graspness, zeng2022robotic}, retrieving~\cite{xu2023joint, huang2021visual}, and rearranging~\cite{goyal2022ifor, cheong2020relocate}, yet it remains highly challenging due to severe occlusions, tight object arrangements, and unstable configurations. 
Previous works have addressed these difficulties through methods such as Visual-Language-Action Model (VLA)~\cite{deng2025graspvla}, multimodal reasoning~\cite{qian2024thinkgrasp}, and synthetic simulation data~\cite{chen2025robotwin}. However, These works primarily focus on cluttered rigid object scenarios without considering cluttered deformable object scenarios.
GarmentPile~\cite{wu2025garmentpile} represents the first attempt at garment manipulation in cluttered environments, but suffers from low compatibility with downstream tasks, single-arm constraints, and lack of language guidance. Our GarmentPile++ addresses these limitations by integrating VLM-based understanding with dual-arm coordination for robust garment manipulation in cluttered scenarios.

\begin{figure*}[!ht]
\vspace{-2mm}
\centering
\includegraphics[width=1.0\linewidth]{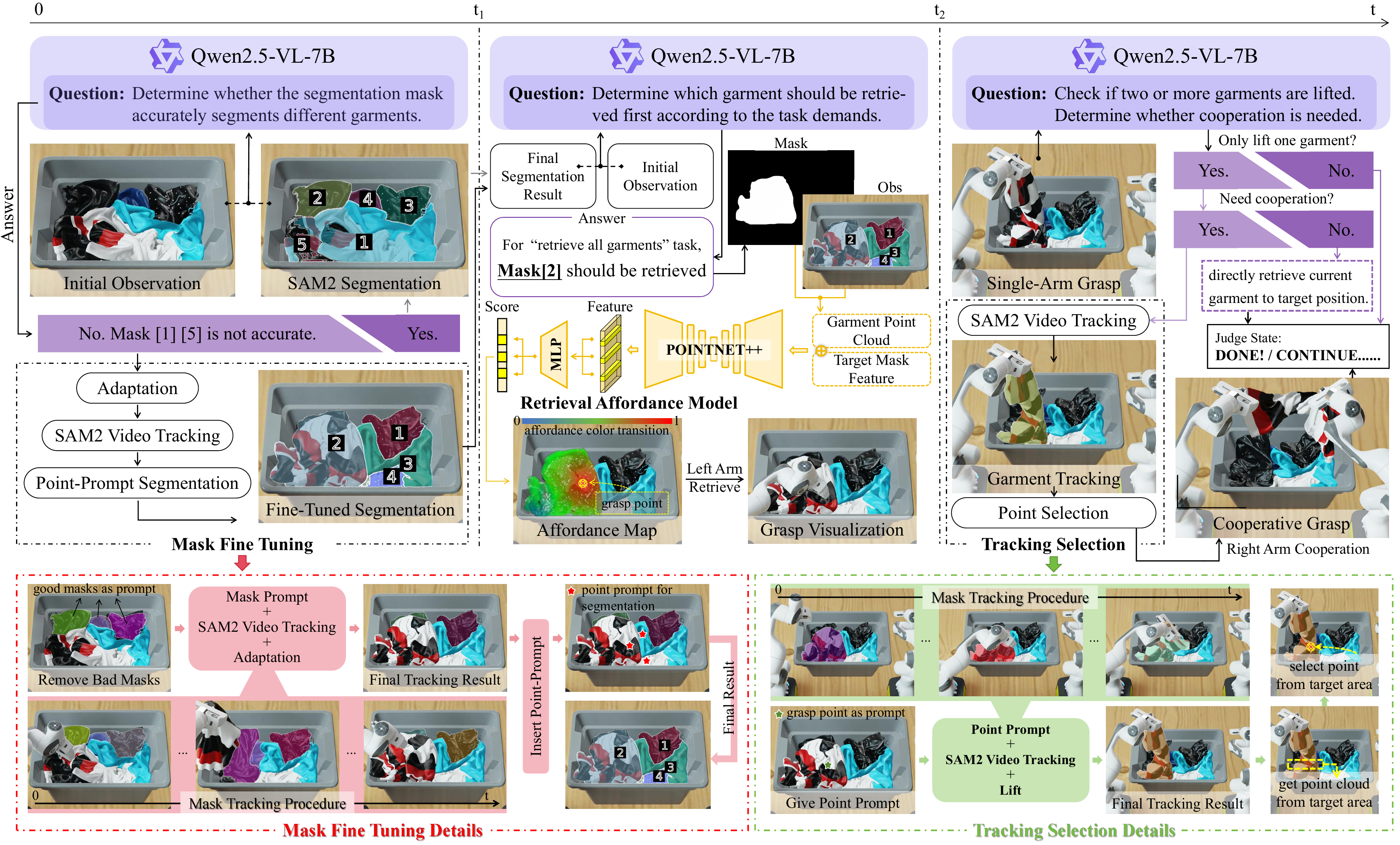}
\vspace{-6mm}
\caption{\textbf{GarmentPile++ Pipeline.} Given the initial observation, SAM2 segments the garment pile. Due to pile complexity, masks may be inaccurate, thus optionally triggering mask fine tuning procedure (Bottom-Left, red) based on VLM reasoning. The finalized mask and observation are then passed to the VLM to select the target garment under task constraints. Treating the target mask as garment pile point-cloud feature, the Retrieval Affordance Model predicts single-arm grasp points, which is used for single-arm retrieval. Then the post-grasp observation is re-queried by the VLM to decide on dual-arm cooperation; if required, a tracking-selection module (Bottom-Right, green) produces cooperative grasp points to ensure successful retrieval. What's more, if VLM detects the single-arm grasp of more than one garment, the current retrieval will be terminated.
}
\label{fig:pipeline}
\vspace{-6mm}
\end{figure*}

\section{PROBLEM FORMULATION}

For a cluttered garment pile, at certain time step $t$, given an RGB-D observation $\mathcal{I}_t \in \mathbb{R}^{H \times W \times 3}$, $\mathcal{D}_t \in \mathbb{R}^{H \times W}$ and an open-ended language query $\mathcal{T}$, the goal of \textit{\textbf{GarmentPile++}} is to provide a policy $\pi(a^1_t, a^2_t \mid \mathcal{I}_t, \mathcal{D}_t, \mathcal{T})$, which converts $\mathcal{T}$ into the target grasping object $\mathcal{X}_t$, and generates actions via $(a^1_t, a^2_t) = \mathrm{G}(\mathcal{X}_t, \mathcal{I}_t, \mathcal{D}_t)$. Herein, $a^1_t$ denotes the action of master arm and $a^2_t$ denotes the action of slave arm. Specially, $a^2_t$ can be set to None in scenarios where cooperative operation is not requisite. The policy $\pi$ will retrieve garments one by one in a safe and neat manner until $\mathcal{T}$ is satisfied.

With reference to previous deformable manipulation studies~\cite{wu2023foresightful, wu2025garmentpile}, we adopt pick-and-place as action primitive. Since place positions are typically predefined (e.g., placing garments on the table), we use the grasp point $p_{retrieve} \in \mathbb{R}^3$ with heuristic retrieval orientation as retrieval action $a_t$.

Our tasks consist of two stages:

\begin{enumerate}
    \item identify the target garment $\mathcal{X}_t$ according to $\mathcal{T}$ and generate master-arm grasping action $a^1_t$. 
    \item determine whether dual-arm cooperation is required for $\mathcal{X}_t$ and generate slave-arm cooperation action $a^2_t$.
\end{enumerate}

In stage \MakeUppercase{\romannumeral 1}, we define affordance map $\mathcal{A}_{retrieve}$, each digit normalized to $[0,1]$, indicating per-point actionability of $\mathcal{X}_t$ for retrieval. The point with the highest score is selected as $p_{retrieve}$ for $a^1_t$. In stage \MakeUppercase{\romannumeral 2}, if cooperation is required, the tracking selection procedure will generate $p_{retrieve}$ for $a^2_t$. Otherwise, $a^2_t$ will be set to None.

\section{METHOD}

\subsection{Overview}

\textit{\textbf{GarmentPile++}} fuses the vision-language reasoning capacity of VLMs with pre-trained instance segmentation model and affordance model. The overall pipeline is shown in Fig.~\ref{fig:pipeline}
. We first describe the details of garment segmentation leveraging SAM2 and selection via VLM (Section~\ref{what_to_retrive}). Then we introduce the architectural design of Retrieval Affordance Model, alongside its inference workflow and training procedure (Section~\ref{where_to_retrieve}). Finally, we delve into the details about master-slave arms cooperation mechanism in GarmentPile++ (Section~\ref{how_to_retrieve}).

\subsection{Vision-Language-Guided Segmentation and  Selection}
\label{what_to_retrive}

\textbf{Single-Image Garment Segmentation}. We employ a SAM2-based model $\mathcal{S}_{image}\!:\!\mathbb{R}^{H\times W\times 3}\!\to\!\{0,1\}^{H\times W}$ to perform scene segmentation. It segments all objects in the scene and produces masks. Using the depth map $\mathcal{D}_t$, we filter out implausible masks (e.g., overly fragmented or nearly planar) to obtain the final set $M_{1:N'}$. For each $M_i$, we choose a location on the mask to place a numeric visual marker $i$ and render annotated images $\mathcal{I}_t^{border,label},\mathcal{I}_t^{mask,label}$ (Border refers to the border of the mask, drawn with a white line.). The annotated images, together with the VLM's OCR capability, enable the VLM to refer to each region by its ID.

\textbf{Mask-Adjustment Decision}. To mitigate SAM2's limitations with garment pile segmentation caused by heavy occlusion and color similarity (e.g., several garments fused into one mask, ragged boundaries), we ask the VLM to determine whether any adjustment is required and, if so, which masks are in need of adjustment. Let the VLM decision function be $\mathcal{F}^{adjust}$, such that $\{I\mid I\subseteq [1{:}N']\}\!\ni\!\mathcal{A}^{adjust}\!=\!\mathcal{F}^{adjust}\!\left(\mathcal{I}_t^{border,label},\mathcal{I}_t^{mask,label}\right)$ where $i\!\in\!\mathcal{A}^{adjust}$ means mask $M_i$ is in need of adjustment.

\begin{figure*}[htbp]
\vspace{-2mm}
\centering
\includegraphics[width=1.0\linewidth]{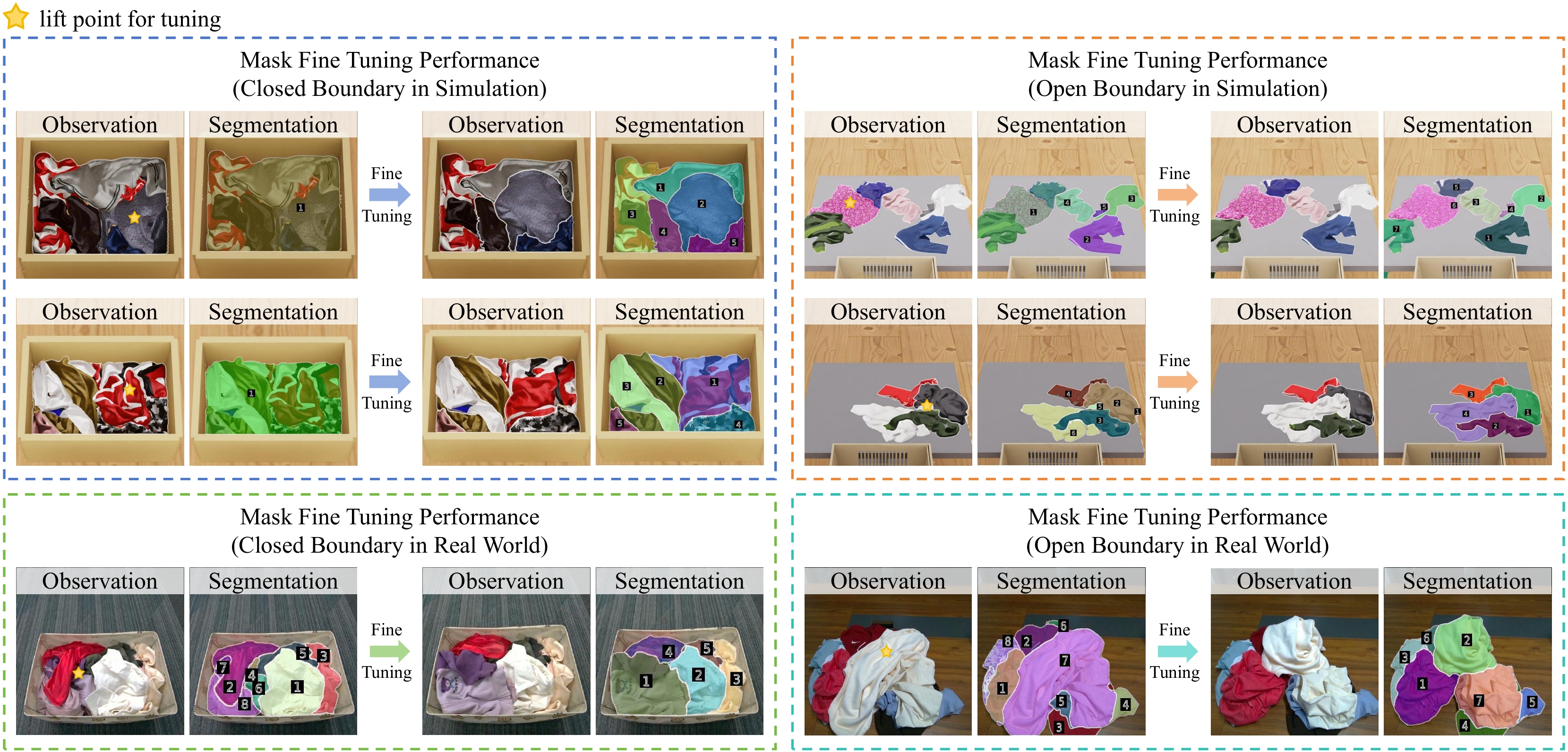}
\caption{\textbf{Segmentation Performance before and after Mask Fine Tuning Procedure.} Initial segmentation result (left-side images of each part) may be flawed : e.g., a mask covering multiple garments (mask 1 in blue part row 1, 2; mask 1 in orange part row 1), fragmented masks (mask 5 in orange part row 2; mask 4,6,7 in green part), or color confusion (mask 7 in cyan part covering two white garments). \ding{73} denotes the lift point for tuning. After fine tuning, the adjusted masks (right-side images of each part) are clearly more reasonable, with each mask generally corresponding to a single complete garment. }
\label{fig:mask_fine_tuning}
% \vspace{-2mm}
\end{figure*}

\begin{figure*}[htbp]
\centering
\vspace{-3mm}
\includegraphics[width=0.95\linewidth]{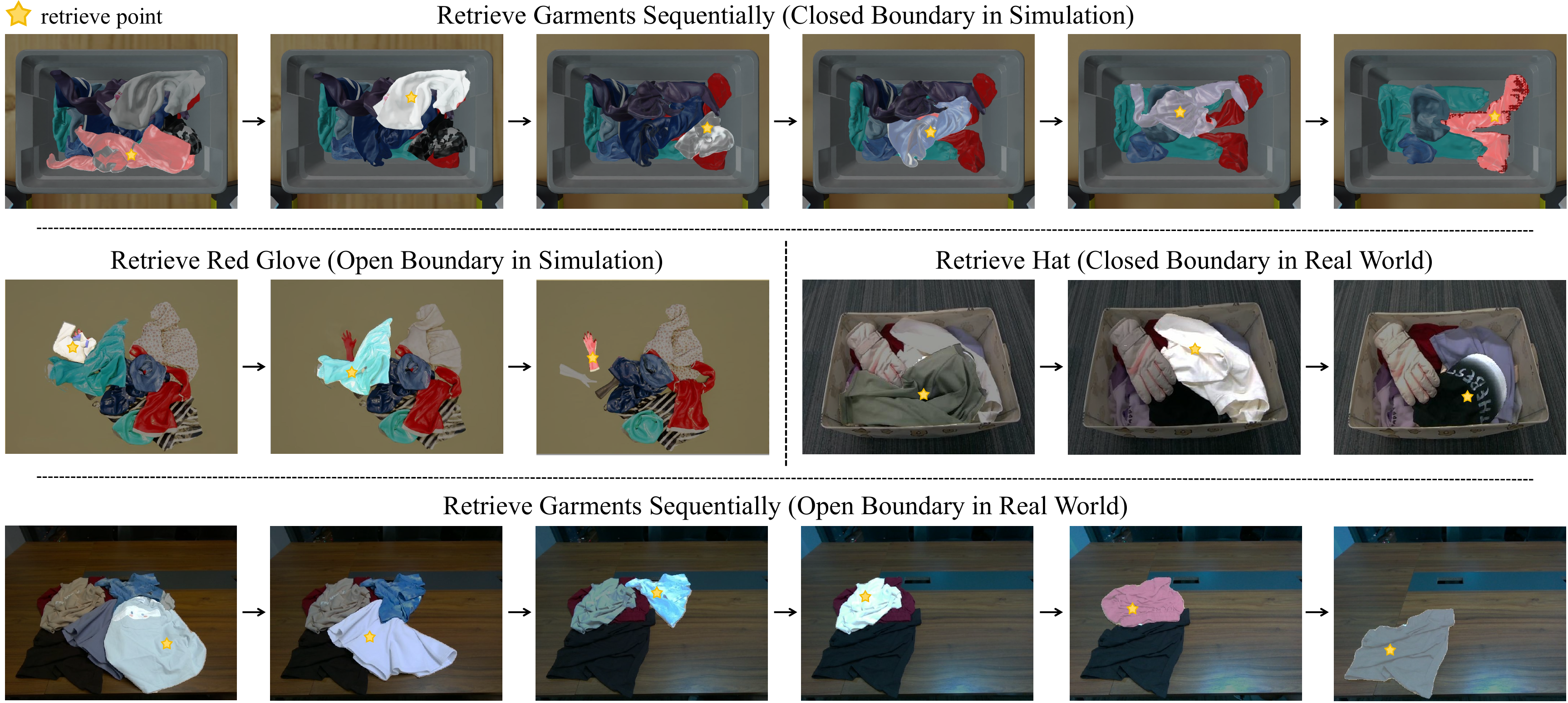}
\caption{\textbf{Reasoning Procedure.} For sequential garment retrieval (rows 1 and 3), the VLM infers the most suitable garment to retrieve (e.g., topmost or with simpler entanglement) to maximize overall success. For specific garment retrieval (row 2), it inters to remove obstructing garments so that the target garment is quickly exposed and ready for retrieving.}
\label{fig:reasoning}
\vspace{-7mm}
\end{figure*}

\textbf{Mask Fine Tuning}. If $\mathcal{A}^{adjust}\neq\emptyset$, We command the robot to pinch the garment at a random point within $\bigcup_{i\in \mathcal{A}^{adjust}}M_i$, lift, shake, and release the garment while recording a video. Using SAM2's \emph{VideoPredictor}, we track $M_{\,\{1{:}N'\}\setminus \mathcal{A}^{adjust}}$ over the sequence, then regenerate masks on the last frame as follows: we start from the image center and treat any pixel not covered by a mask as a point prompt for SAM2 to generate a new mask. After filtering out implausible masks and applying Non-Maximum Suppression (NMS), we obtain the final set $M_{1:N}$ and corresponding numeric markers $i$. Fig.~\ref{fig:mask_fine_tuning} shows the performance of mask fine tuning procedure.

\textbf{Vision–Language–Guided Selection}. Let the VLM selection function be $\mathcal{F}^{selection}$, such that $\{1{:}N\}\!\ni\!n^{selection}\!=\!\mathcal{F}^{selection}\!\left(\mathcal{I}_t^{border,label},\mathcal{T}\right)$
. Hence, $M_{n^{selection}}$ is the mask of the target garment $\mathcal{X}_t$. Fig.~\ref{fig:reasoning} illustrates the reasonability of VLM-based reasoning.

\subsection{Affordance-Guided Garment Retrieval}
\label{where_to_retrieve}

\begin{figure}[htbp]
\vspace{-2mm}
\centering
\includegraphics[width=1.0\linewidth]{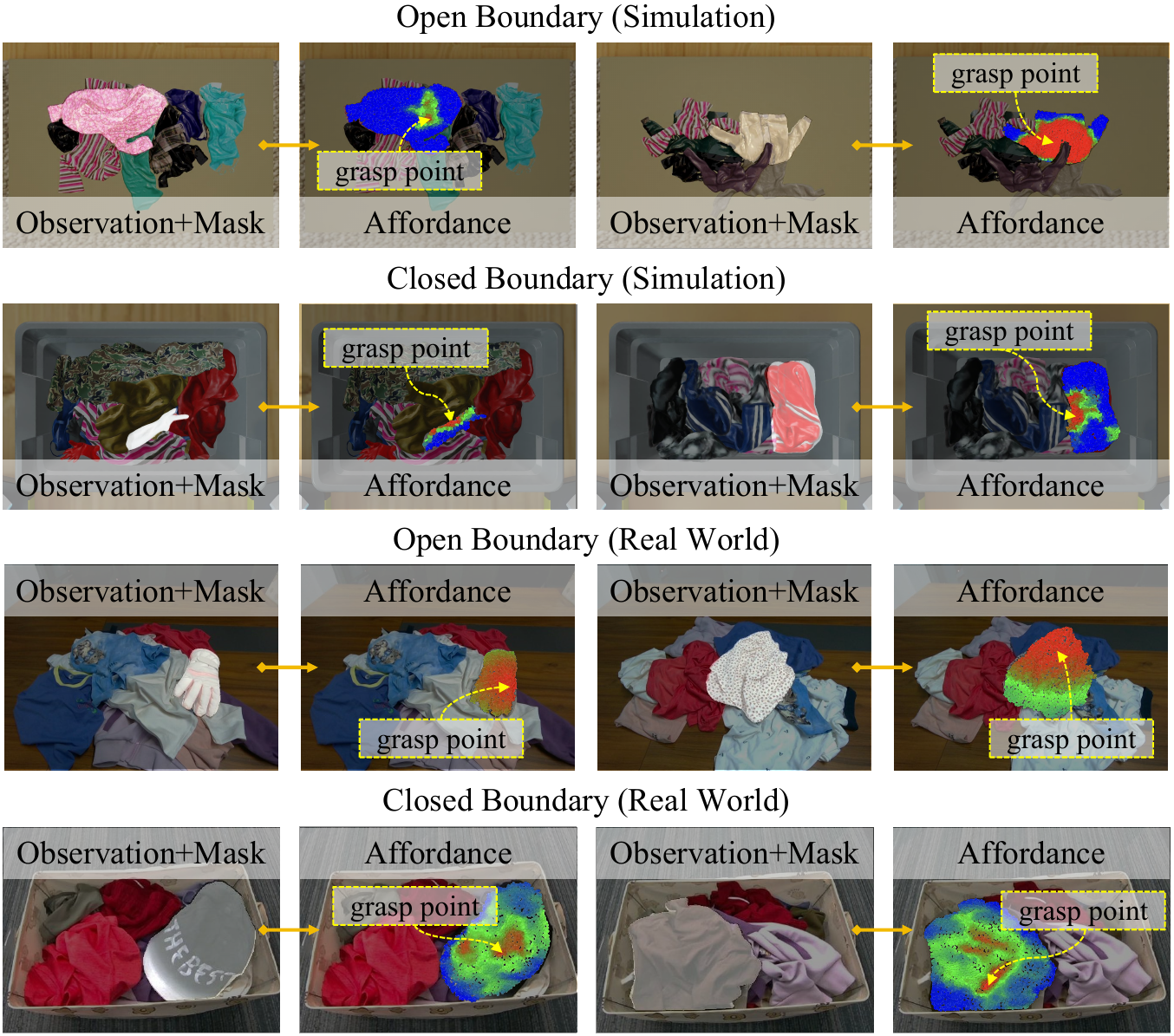}
\caption{\textbf{Affordance Visualization.} The color transition (blue $\rightarrow$ green $\rightarrow$ red) indicates an increasing affordance level. Beyond generalization across different garment types, the affordance further captures: \textbf{garment geometry} (row 1, center regions of garments are typically preferable.), \textbf{subtle structures} (row 4, garment wrinkle regions easier to grasp exhibit higher affordance level.), \textbf{spatial relations} (row 3 left, grasping left / right region of glove is equivalent, but the right side is preferable due to less garments, as well as tops side of garment in row 3 right; row 2, central regions far from closed boundaries are preferable for grasp safety.)}
\label{fig:affordance}
\vspace{-6mm}
\end{figure}

The $\mathcal{M}_{\text{aff}}:(\mathcal{O}, M_{1:N}, n^{selection})\!\to\![0,1]^{\mathcal{O}}\!=\!\mathcal{A}^{retrieve}$ is our Retrieval Affordance Model, which takes the point-cloud observation $\mathcal{O}$ of the garment clutter and the partial results in Section~\ref{what_to_retrive} as input and produces the affordance map $\mathcal{A}^{retrieve}$ indicating the per-point actionability of $\mathcal{X}_t$ for retrieval with a single arm, as shown in Fig.~\ref{fig:affordance}.

Since $\mathcal{O}$ is projected from $\mathcal{I}_t$, we denote by $\mathcal{O}^{M_i}$ the point-cloud observation on $\mathcal{I}_t$ covered by mask $M_i$. $\mathcal{M}_{\text{aff}}$ first employs the PointNet++~\cite{qi2017pointnet++} backbone feature extractor $\mathbf{F}_{retrieve}$. It takes as input the 3D coordinates in $\mathcal{O}^{M_{1:N}}$ with the 1-dimensional feature $[1\ \textrm{if}\ p\in \mathcal{O}^{M_{n^{selection}}}\ \mathrm{else}\ 0]$ and extracts the feature $f_p^{retrieve}\in\mathbb{R}^{128}$ for each point $p$ in $\mathcal{O}^{M_{1:N}}$. It then feeds these features into multi-layer perceptrons (MLPs) and, after normalization with sigmoid activation, yields the 1-dimensional affordance prediction $\hat{g}_p^{retrieve}\in[0,1]$ for $p$. the point $p$ with highest score on $\mathcal{O}^{M_{n^{selection}}}$ is selected as retrieval point $p_{retrieve}$.

As for training procedure, We denote by $g_p^{retrieve}\in\{0,1\}$ the ground-truth retrieval affordance score on $p$. We directly perform single-arm grasping at $p$, and assign $g_p^{retrieve}$ to 1 if the grasp-and-place succeeds, and to 0 if it fails. The loss is computed using Binary Cross Entropy(BCE) Loss~\cite{ruby2020binary}, i.e.,
\begin{align}
    L_{retrieve}
      = & - \Big(g_p^{retrieve} \cdot \log(\hat{g}_p^{retrieve}) \notag \\
      & + (1 - g_p^{retrieve}) \cdot \log(1 - \hat{g}_p^{retrieve})\Big).
\end{align}
The points $p$ participating in training satisfy $p\in\mathcal{O}^{M_{n^{selection}}}$, since ultimately only the highest-scoring point in $\mathcal{O}^{M_{n^{selection}}}$ is selected for the retrieval action. By retaining the remaining points which belong to $\mathcal{O}$ but not to $\mathcal{O}^{M_{n^{selection}}}$, the point-level features extracted by PointNet++ can not only capture the local structure of $\mathcal{O}^{M_{n^{selection}}}$ but also aggregate information about the global structure and inter-garment relations. Such information is crucial for predicting $\hat{g}_p^{retrieve}$.

\subsection{State-Based Arm Cooperation}
As mentioned in~\ref{where_to_retrieve}, $p_{retrieve}$, the highest-scoring point on $\mathcal{O}^{M_{n^{selection}}}$, is selected for retrieval action. According to the position of $p_{retrieve}$ in the scene, we choose the arm on the closer side as the master arm and use it to lift the garment at $p_{retrieve}$.

We record the lifting process on video and employ SAM2’s Video Predictor to track $M^{selection}$ from start to end until $M^{picked}$. This avoids redundant re-segmentation of the scene and prevents other garments with similar colors from being mistakenly merged into the same mask in the post-lifting 2D image.

Next, we assign the numeric visual marker $1$ to mask $M^{picked}$. The VLM’s cooperation function $\mathcal{F}^{cooperation}$ then determines whether two or more garments are lifted and whether master-slave arms cooperation is required (e.g., considering whether the garment is too long), i.e., 
$\textstyle (\{0,1\} \ni x^{error}, \{0,1\} \ni x^{dual}) = \mathcal{F}^{cooperation}(\mathcal{I}_{t}^{border,label}, \mathcal{I}_{t'}^{border,label})$
, where $t$ is the time step defined in Section~\ref{what_to_retrive}, and $t'$ denotes the post-lifting time step. In this setting, we prompt the VLM to align two garments in the scene to better identify the lifted garment.

$x^{error}=1$ means two or more garments are lifted, which results in termination of current retrieval and begin of next attempt. When $x^{error}=0$, only one garment is lifted, allowing the cooperation process to proceed. If $x^{dual}=0$, no master-slave arms cooperation is required; the master-arm directly delivers the garment to the target position. Otherwise, we use the function $\mathcal{P}^{dual}:(\mathcal{D}_{t'}, M^{picked})\to p^{dual}$ to select the grasping point of the slave-arm. Specifically, all points in $M^{picked}$ are sorted along the $z$-axis, and a pick point is sampled from the bottom. Finally, the slave-arm grasps the selected point to lift, and then both arms cooperatively deliver the garment horizontally to the target location.

\label{how_to_retrieve}

\section{EXPERIMENT}

\subsection{Simulation Experiments}

\begin{table*}[htbp]
\vspace{-2mm}
\centering
\setlength{\tabcolsep}{4pt}
\caption{Comparison of baselines on garment retrieval tasks.}
\label{tab:baseline}
\begin{tabular*}{\textwidth}{@{\extracolsep{\fill}} c c c c c c c c c c c c c @{}}
\toprule[1.2pt]
\multirow{3}{*}{\textbf{Method}} & 
\multicolumn{4}{c}{\textbf{Retrieve Garments Sequentially}} & 
\multicolumn{8}{c}{\textbf{Retrieve a Piece of Specific Garment}} \\ 
\cmidrule(lr){2-5} \cmidrule(lr){6-13}
& \multicolumn{2}{c}{Open Boundary} & \multicolumn{2}{c}{Closed Boundary} 
& \multicolumn{2}{c}{\begin{tabular}{c} Hat \\ (Open Boundary) \end{tabular}} 
& \multicolumn{2}{c}{\begin{tabular}{c} Glove \\ (Closed Boundary) \end{tabular}} 
& \multicolumn{2}{c}{\begin{tabular}{c} Green Tops \\ (Open Boundary) \end{tabular}} 
& \multicolumn{2}{c}{\begin{tabular}{c} Black Trousers \\ (Closed Boundary) \end{tabular}} \\ 
\cmidrule(lr){2-3} \cmidrule(lr){4-5} \cmidrule(lr){6-7} \cmidrule(lr){8-9} \cmidrule(lr){10-11} \cmidrule(lr){12-13}
& \textbf{$\text{ASR}_{\text{{A}}}$$\uparrow$} & \textbf{AMS} & \textbf{$\text{ASR}_{\text{{A}}}$$\uparrow$} & \textbf{AMS} 
& \textbf{$\text{ASR}_{\text{{B}}}$$\uparrow$} & \textbf{AMS} & \textbf{$\text{ASR}_{\text{{B}}}$$\uparrow$} & \textbf{AMS} 
& \textbf{$\text{ASR}_{\text{{B}}}$$\uparrow$} & \textbf{AMS} & \textbf{$\text{ASR}_{\text{{B}}}$$\uparrow$} & \textbf{AMS} \\ 
\midrule
ThinkGrasp   & 0.641 & $\diagdown$ & 0.413 & $\diagdown$ & 0.517 & 2.017 & 0.593 & 1.439 & 0.542 & 1.067 & 0.585 & 1.171 \\ 
GarmentPile   &  0.819 & $\diagdown$ &  0.792 & $\diagdown$ & 0.623 & 2.798 & 0.659 & 3.152 & 0.729 & 1.957 & 0.545 & 1.394 \\ 
Qwen (only)   &  0.829& $\diagdown$ &  0.720& $\diagdown$ & 0.708 & 1.858 &  0.739 &  1.797& 0.787 & 1.160 & 0.686 & 1.081 \\ 
Our Method    & \textbf{0.904} & $\diagdown$ & \textbf{0.874} & $\diagdown$ & \textbf{0.892} & 2.196 & \textbf{0.925} & 2.142 &  \textbf{0.877} & 1.106 & \textbf{0.862} & 1.147 \\ 
\bottomrule[1.2pt]
\end{tabular*}
\vspace{-4mm}
\end{table*}

\textbf{Environment Setup.} 
We use DexGarmentLab~\cite{wang2025dexgarment} built on Isaac Sim 4.5.0~\cite{NVIDIA_Isaac_Sim} as our backbone simulation environment, which supports both multi-garments interactions and garment-robot interactions. We construct two different types of representative scenario for garment retrieval:
\begin{itemize}
    \item \noindent\textit{Open-Boundary Scenario:} In this scenario, garments are randomly scattered across relatively open planar surfaces (e.g., tables, floors, beds, etc.). This implies that the stacking state of garment piles is relatively insignificant, as spatial constraints are absent.
    \item \noindent\textit{Closed-Boundary Scenario:} In this scenario, garments are randomly placed into containers with enclosed boundaries (e.g., baskets, boxes, etc.). This implies that, due to spatial constraints, the garment piles will exhibit a relatively severe stacking state.
\end{itemize}

We load 9 categories (dress, trousers, tops, skirt, socks, glove, hat, scarf and underpants) of 153 different garments from ClothesNet~\cite{zhou2023clothesnet}. In the Open-Boundary Scenario, 6–16 garments will be loaded to form garment piles; in the Closed-Boundary Scenario, the quantity of garments employed will be 3–8. What's more, there are two types of typical tasks conducted in each scenario:

\begin{itemize}
    \item \noindent\textit{Task A: Retrieve Garments Sequentially.} The robot is required to continue retrieving until no garments remain in the observation field.
    \item \noindent\textit{Task B: Retrieve a Piece of Specific Garment.} The robot needs to retrieve until the target garment (e.g., hat, glove, green tops, black trousers, etc.) is retrieved. 
\end{itemize}

\textbf{Baselines.}
Owing to the lack of relevant studies in cluttered garments retrieval with VLM reasoning, we select representative works in the fields of VLM-guided grasping and cluttered garment retrieval, and adapt them to align with the requirements of our tasks. The baselines and their introductions are shown as below: (1) \textbf{ThinkGrasp}~\cite{qian2024thinkgrasp}. According to the current RGB observation, the modified ThinkGrasp first give a brief introduction about the target garment, which is then used by LangSAM to segment and output the mask. The VLM select a grasp point inside the mask and guide the single arm to grasp and retrieve the garment. (2) \textbf{GarmentPile}~\cite{wu2025garmentpile}. The robot continues retrieving garment only according to the point cloud of garment piles without VLM reasoning, until the task is finished. (3) \textbf{Qwen (only)}~\cite{bai2025qwen}. Qwen2.5-VL-7B is used as the same with our method in this baseline. It first infers an appropriate grasping point based on the given RGB observation, and then guides the single arm to grasp and lift the garment. Subsequently, the current RGB observation is captured and sent to Qwen2.5-VL-7B to determine whether to use the dual arms and which point to select for cooperation grasp.

We evaluate performance using Average Success Rate (ASR) for both tasks, with Average Motion Steps (AMS) as an additional efficiency metric for Task B.

\noindent\textit{Average Success Rate (ASR):} For Task A, ASR measures the probability of successfully grasping any single garment:
\begin{equation}
    \text{ASR}_{\text{A}} = \frac{\text{Number of successfully grasped garments}}{\text{Total number of loaded garments}}
\end{equation}

For Task B, ASR represents the probability of successfully retrieving the target garment while maintaining proper handling of all previously grasped items:
\begin{equation}
    \text{ASR}_{\text{B}} = \frac{\text{Number of completed tasks}}{\text{Total number of tasks}}
\end{equation}

\noindent\textit{Average Motion Steps (AMS):} This metric quantifies the efficiency of our approach by measuring the average number of motion steps required to complete each task:
\begin{equation}
    \text{AMS} = \frac{\text{Total motion steps to complete all tasks}}{\text{Total number of tasks}}
\end{equation}

\textbf{Ablations.}
To demonstrate the necessity of the proposed adaptation module, we compare with the following ablated versions:
(1) \textbf{w/o Mask Fine Tuning} that removes the mask fine-tuning part and directly uses masks generated by SAM2 scene segmentation. (2) \textbf{w/o Affordance} that removes the affordance part and randomly selects grasp point from target garment mask. (3) \textbf{w/o Tracking Selection} that removes the tracking selection part and directly queries Qwen2.5-VL-7B using single-arm lift image to select cooperative point for dual arm. (4) \textbf{w/o Dual Arm} that removes dual-arm cooperation part and only uses single for garment retrieval. (5) \textbf{w/o Affordance \& Dual Arm} that removes both affordance and dual-arm cooperation part and randomly selects grasp point for single arm to retrieve garment.

For all the ablations, the ASR metric is used for evaluation; while for w/o Affordance baseline and w/o Mask Fine Tuning baseline, the Probability of Dual Arm (PDR) metric is additionly used for evaluating the probability of triggering dual-arm collaborative operations, which reprensents the efficiency of pipeline.

\textbf{Results and Analysis.} Tab.~\ref{tab:baseline} and Tab.~\ref{tab:ablation} present the baseline comparison results and ablation study results for our method, correspondingly. Fig.~\ref{fig:mask_fine_tuning}, Fig.~\ref{fig:reasoning} and Fig.~\ref{fig:affordance} respectively show the visualization performance of the mask fine-tuning part, reasoning procedure part and affordance part in our proposed method.

Tab.~\ref{tab:baseline} demonstrates that our method outperforms all baselines.
For \textbf{ThinkGrasp}, due to the complex states and diverse colors of garments, it is challenging to guide LangSAM to obtain appropriate and accurate masks via linguistic descriptions. Additionally, ThinkGrasp only adopts single-arm grasping, resulting in relatively low success rates across all tasks. 
For \textbf{GarmentPile}, it leverages affordance to identify the most accessible single-arm grasping position on the garment pile. In terms of success rate, it generally outperforms ThinkGrasp. However, the lack of linguistic guidance leads to no clear direction when determining garment grasping positions, resulting in a significantly higher AMS than other methods and thus lower efficiency. 
As for \textbf{Qwen (only)}, it incorporates both VLM reasoning and dual-arm cooperation modules, contributing to a better success rate than ThinkGrasp and GarmentPile. Nevertheless, it is highly affected by the garment stacking state: in the Open Boundary scenario, where stacking is mild and garment structures are relatively distinct, Qwen achieves a high success rate; in the Closed Boundary scenario, severe garment stacking limits its reasoning capability, leading to a marked drop in success rate.
\textbf{Our method} integrates multiple components, including mask fine-tuning, affordance, VLM reasoning, and dual-arm cooperation. It achieves state-of-the-art (SOTA) performance for cluttered garment manipulation across all task scenarios and its AMS shows slight differences from other baselines, ensuring both high success rate and efficiency.

\begin{table}[tbp]
  \vspace{-2mm}
  \centering
  \setlength{\tabcolsep}{4pt}
  \caption{Ablation of our methods on two typical tasks}
  \label{tab:ablation}
  \resizebox{0.495\textwidth}{!}{
  \begin{tabular}{lcccc}
    \toprule
    \textbf{Method} & \multicolumn{2}{c}{\makecell{Retrieve a Piece of \\ Green Tops \\ (Open Boundary)}} & \multicolumn{2}{c}{\makecell{Retrieve Garments \\ Sequentially \\ (Closed Boundary)}} \\
    \cmidrule(lr){2-3} \cmidrule(lr){4-5}
    & \multicolumn{2}{c}{\textbf{$\text{ASR}_{\text{{B}}}$ $\uparrow$}} & \multicolumn{2}{c}{\textbf{$\text{ASR}_{\text{{A}}}$ $\uparrow$}} \\
    % \makecell[l]{w/o Mask Fine Tuning} & \multicolumn{2}{c}{0.841} & \multicolumn{2}{c}{} \\
    \makecell[l]{w/o Tracking Selection} & \multicolumn{2}{c}{0.669} & \multicolumn{2}{c}{0.739} \\
    \makecell[l]{w/o Dual Arm} & \multicolumn{2}{c}{0.613} & \multicolumn{2}{c}{0.634} \\
    \makecell[l]{w/o Affordance \& Dual Arm} &  \multicolumn{2}{c}{0.543} & \multicolumn{2}{c}{0.564} \\
    \cmidrule(lr){2-3} \cmidrule(lr){4-5}
    & \textbf{$\text{ASR}_{\text{{B}}}$ $\uparrow$} & \textbf{PDR $\downarrow$} & \textbf{$\text{ASR}_{\text{{A}}}$ $\uparrow$} & \textbf{PDR $\downarrow$} \\
    \makecell[l]{w/o Mask Fine Tuning} & 0.841 & 0.794 &  0.857 & 0.624 \\
    \makecell[l]{w/o Affordance} & 0.823 & 0.843 & 0.810 & 0.595 \\
    \makecell[l]{Our Method} & \textbf{0.877} & \textbf{0.766} & \textbf{0.874} & \textbf{0.551} \\
    \bottomrule
  \end{tabular}
  }
  \vspace{-6mm}
\end{table}

Table~\ref{tab:ablation} presents a quantitative analysis of the ablation study, demonstrating the effectiveness of key components in our method, including \textit{mask fine tuning}, \textit{tracking selection}, \textit{dual-arm cooperation}, and \textit{affordance}.
Notably: 
1) The comparison between the "w/o Dual Arm" and "w/o Affordance \& Dual Arm" reveals that integrating the affordance component enhances the success rate of single-arm grasping. This demonstrates that the affordance guides the robotic arm to prioritize manipulating the more suitiable region of garments, preventing excessive sagging.
2) Based on this observation, the further comparison between "w/o Affordance" and "Our Method" (in terms of ASR and PDR) shows that the affordance component not only improves the overall success rate but also reduces the probability of triggering dual-arm cooperation, which boosts efficiency.
3) Besides, the comparison between "w/o Mask Fine Tuning" and "Our Method" (in terms of ASR and PDR) establishes that our mask fine tuning procedure enhances the accuracy of segmented masks, which enables the affordance component to locate better points for retrieval according to the precise garment mask, ultimately leading to higher ASR score and lower PDR score.
4) Additionally, other results also indicate that dual-arm cooperation and tracking selection significantly elevate the overall task success rate, which further underscores the necessity of adopting dual-arm cooperation for cluttered garments manipulation, and exposes the deficiency in the understanding and reasoning capabilities of current VLMs for garments without auxiliary information cues.

Fig.~\ref{fig:mask_fine_tuning}, Fig.~\ref{fig:reasoning} and Fig.~\ref{fig:affordance} further illustrate the effectiveness of \textit{mask fine tuning}, \textit{VLM reasoning} and \textit{affordance}. In Fig.~\ref{fig:mask_fine_tuning}, the adjusted mask enables more accurate segmentation of individual garments in the scene, thereby supporting the subsequent reasoning and judgment of the VLM. In previous works~\cite{qian2024thinkgrasp, jiao2025free}, since the task scenarios only involved rigid objects where segmentation is relatively less challenging, no additional mask adjustment was required. However, in scenarios with cluttered garments, accurate segmentation of garments cannot be guaranteed, which necessitates the adoption of the mask fine-tuning method.
In Fig.~\ref{fig:reasoning}, it is evident that the VLM can reasonably infer the mask of the garment to be grasped based on task requirements, which further validates the rationality of the overall system.
In Fig.~\ref{fig:affordance}, the visualized affordance clearly demonstrates its generalization ability and awareness of garment structure and relations. This not only improves the manipulation success rate but also reduces the probability of triggering dual-arm cooperation, thereby enhancing efficiency.

\subsection{Real-World Experiments}

\begin{table}[!t]
  \vspace{-2mm}
  \centering
  \setlength{\tabcolsep}{4pt}        % 调整列间距
  \caption{Real-World Performance Comparison}
  \resizebox{0.495\textwidth}{!}{
  \begin{tabular}{c c c c c}
    \toprule
    \textbf{Real-World Task} & Metrics & ThinkGrasp & Qwen(only) & Our Method \\
    \midrule
    \multirow{2}{*}{\makecell{Retrieve Garments Sequentially \\ (Open Boundary)}} & 
    \multirow{2}{*}{\textbf{$\text{ASR}_{\text{{A}}}$ $\uparrow$}} & \multirow{2}{*}{9/20 (45\%)} & \multirow{2}{*}{16/20 (80\%)} & \multirow{2}{*}{\textbf{19/20 (95\%)}} \\
    & & & \\
    \cmidrule(lr){2-2} 
    \multirow{2}{*}{\makecell{Retrieve Garments Sequentially \\ (Closed Boundary)}} & 
    \multirow{2}{*}{\textbf{$\text{ASR}_{\text{{A}}}$ $\uparrow$}} & \multirow{2}{*}{7/20 (35\%)} & \multirow{2}{*}{11/20 (55\%)} & \multirow{2}{*}{\textbf{17/20 (85\%)}} \\
    & & & \\
    \cmidrule(lr){2-2} 
    \multirow{2}{*}{\makecell{Retrieve a Piece of Hat \\ (Open Boundary)}} & 
    \multirow{2}{*}{\textbf{$\text{ASR}_{\text{{B}}}$ $\uparrow$}} & \multirow{2}{*}{5/10 (50\%)} & \multirow{2}{*}{7/10 (70\%)} & \multirow{2}{*}{\textbf{9/10 (90\%})} \\
    & & & \\
    \cmidrule(lr){2-2} 
    \multirow{2}{*}{\makecell{Retrieve a Piece of Purple Tops \\ (Closed Boundary)}} & 
    \multirow{2}{*}{\textbf{$\text{ASR}_{\text{{B}}}$ $\uparrow$}} & \multirow{2}{*}{4/10 (40\%)} & \multirow{2}{*}{5/10 (50\%)} & \multirow{2}{*}{\textbf{8/10 (80\%})} \\
    & & & \\
    \bottomrule
  \end{tabular}
  }
  \label{tab:real_world_performance}
  \vspace{-6mm}
\end{table}

\begin{figure}[htbp]
\vspace{-4mm}
\centering
\includegraphics[width=1.0\linewidth]{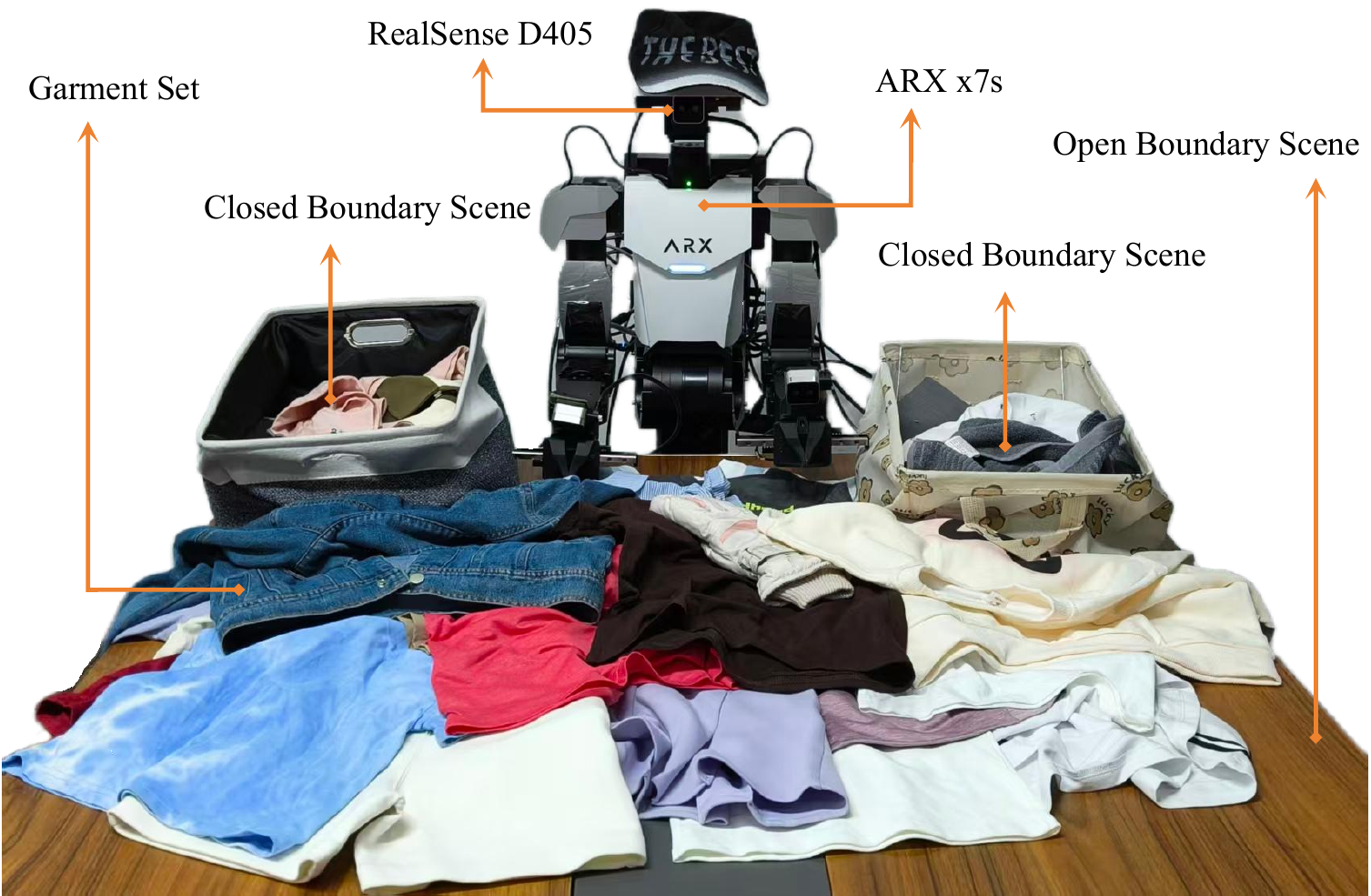}
\caption{\textbf{Real-World Experiments Setting.}}
\label{fig:real_world}
\vspace{-2mm}
\end{figure}

We use ARX x7s as real-world robot and RealSense D405 as head camera for images capture. The real-world experiments setting are shown in Fig.~\ref{fig:real_world}. We have already provided some real-world performances on \textit{mask fine tuning}, \textit{VLM reasoning} and \textit{affordance} in Fig.~\ref{fig:mask_fine_tuning}, Fig.~\ref{fig:reasoning} and Fig.~\ref{fig:affordance} to illustrate the effectiveness of our method. Tab.~\ref{tab:real_world_performance} reports ASR on baselines and our method, which further demonstrates the excellent performance of GarmentPile++. Supplementary video and project website show more demonstrations.

\section{CONCLUSIONS}

we propose a novel cluttered garment retrieval pipeline that fully utilizes VLM's powerful capability of reasoning and strong generalization ability of affordance to retrieve garment one by one based on language instructions, which establish robust foundation for the execution of downstream tasks (e.g., folding, hanging, wearing). 
As our method relies on visual segmentation model, it may suffer from performance degradation under low-light conditions or for garments with complex patterns/colors due to segmentation errors. Nevertheless, it still contributes significantly to building garment manipulation pipeline in home environments.

\vspace{-1mm}
\section*{Acknowledgment}
\vspace{-1mm}

This project was supported by National Natural Science Foundation of China (62376006) and National Youth Talent Support Program (8200800081).

%%%%%%%%%%%%%%%%%%%%%%%%%%%%%%%%%%%%%%%%%%%%%%%%%%%%%%%%%%%%%%%%%%%%%%%%%%%%%%%%

\bibliographystyle{plain}
\bibliography{main}

\end{document}